\documentclass{article}

 \usepackage[dblblindworkshop, final]{neurips_2026}
\workshoptitle{Linguistic Principles for Foundation Models (LP4FM)}

\usepackage[utf8]{inputenc} 
\usepackage[T1]{fontenc}    
\usepackage{hyperref}       
\usepackage{url}            
\usepackage{booktabs}       
\usepackage{amsfonts}       
\usepackage{nicefrac}       
\usepackage{microtype}      
\usepackage{xcolor}         

\usepackage{graphicx}
\usepackage{booktabs}
\usepackage{multirow}
\usepackage{amsmath}

\usepackage[normalem]{ulem}

\usepackage{forest}
\usepackage{adjustbox}

\forestset{
  parse/.style={
    for tree={
      font=\footnotesize,
      inner sep=2.5pt,
      s sep=10pt,
      l sep=3pt
    }
  }
}

\title{Distinguishing among reasoning levels of GPT 5.5 Pro via a grammar lens benchmark}

\author{%
  Emma Graham \\
  Dartmouth College\\
  Hanover, NH 03755 \\
  \texttt{emma.graham.th@dartmouth.edu} \\
  \And
  Richard Granger \\
  Dartmouth College \\
  Hanover, NH 03755 \\
  \texttt{richard.granger@gmail.com} \\
}

\begin{document}

\maketitle

\begin{abstract}
    Does the level of test-time processing effort affect a language model's computational capability? A formal grammar-lens benchmark method, derived from the theory of computation, generates novel data at specified levels, each successively more computationally powerful than the levels below. Using this data, we tested GPT-5.5, and GPT-5.5-Pro-Medium, -High, and -XHigh, showing incrementally increased acceptance of test data: 16, 40, 43, and 52\%, respectively, illustrating the benchmark's ability to discern fine-grained differences among models' added test-time reasoning levels. Additionally, as test-time compute effort increases, model responses diverge from human responses to the same data. This increasing disagreement may turn out to be broadly characteristic of more capable language models, introducing the potential of a general capability-versus-alignment conflict. 
\end{abstract} 

\section{Introduction}

\begin{figure}
    \centering
    \includegraphics[width=\linewidth]{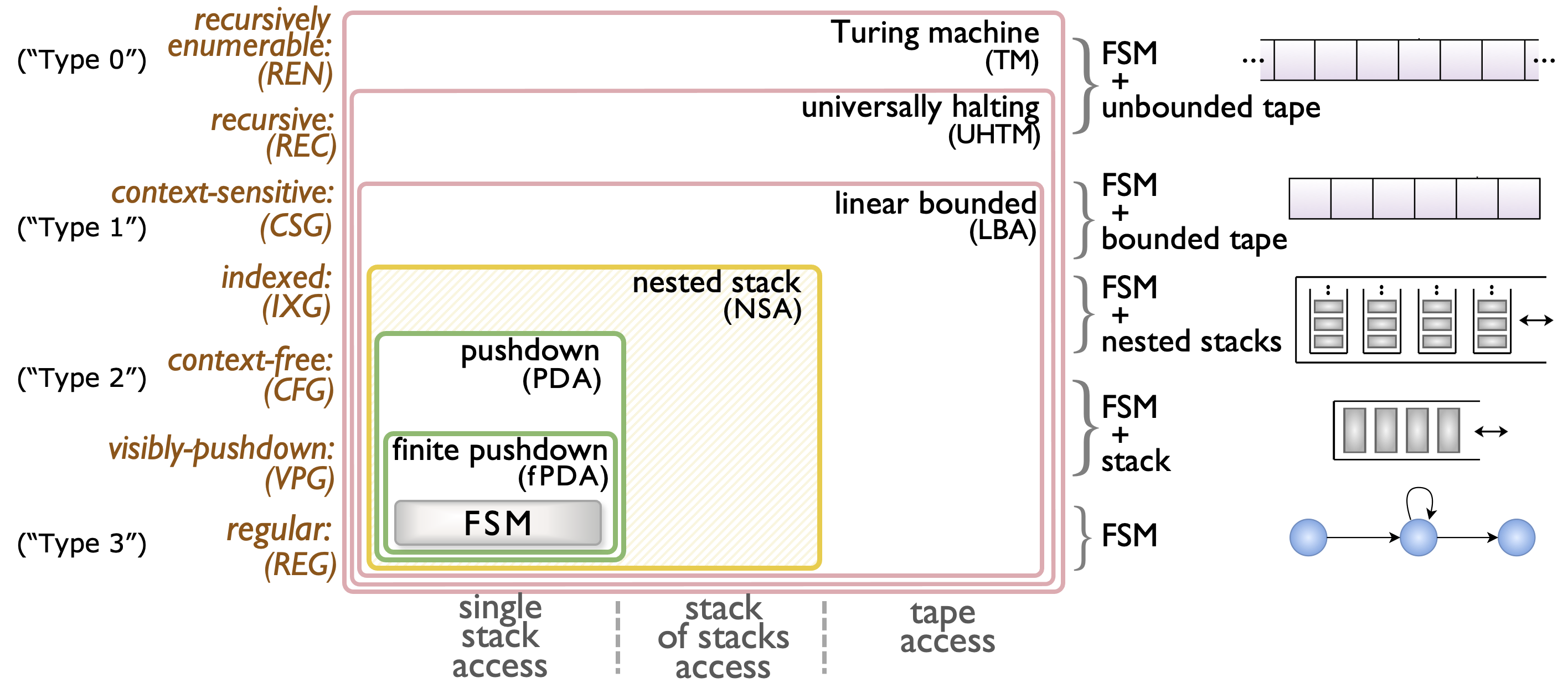}
    \caption{The grammar--automata hierarchy equivalence classes (\emph{left}) with formal grammars (brown) and automata (black) in order of increasing computational power, from finite state machines (FSM) to Turing machines (TM). Single-stack mechanisms (green) are distinguished from higher order of stacks (yellow) and from tape-based automata (red). (\emph{right}) External memory associated with each automata class. (From Graham and Granger~\cite{grahamgranger2026})}
    \label{fig:hierarchy}
\end{figure}

Fundamental abilities and limitations of language foundation models may be analyzed through the lens of computational theory. The theory of computation establishes a hierarchy of formal grammars (which generate sets of strings or \emph{formal languages}) and corresponding automata (which process or recognize those formal languages) (Figure~\ref{fig:hierarchy}). Each tier of the hierarchy defines an equivalence class of computational systems; each tier is computationally equivalent to a finite controller equipped with successively more powerful memory access mechanisms, from pushdown stacks to tapes. Successive tiers are a proper superset of the tiers below it, and each tier enables the processing of increasingly complex relational dependencies \cite{hopcroft2006automata, sipser2006theory, jager2012formal}.

Recent work has started probing large language model capabilities using strings from major tiers of this hierarchy \cite{deletang2023neural, dave2024symbolic, grahamgranger2026}. Most recently, evaluations have generated strings from a context-free grammar (CFG), an indexed grammar (IXG), and a context-sensitive grammar (CSG) with English rules and terminal symbols \cite{grahamgranger2026}. The grammars were used to generate novel, semantically anomalous strings whose dependencies are fixed by one of their three formal grammars, reducing reliance on memorized sentences, world knowledge, and semantic plausibility/probability from model responses. The results indicate a progression from partial CFG string acceptance in smaller models to robust CFG and IXG string acceptance in the largest language models (LLMs). Human participants and conventional LLMs overwhelmingly rejected the CSG-generated strings, whereas some models with augmented inference-time computation (such as chain-of-thought (CoT)\cite{wei2022cot} of mixture-of-experts (MoE) \cite{lo-etal-2025-closer}, mechanisms commonly marketed as ``reasoning'' components), accepted a nontrivial fraction of CSG-generated strings. This raises an interpretive question: when such a model accepts CSG-generated strings, is it making \emph{incorrect} judgments about human natural language syntax, or is it recognizing more complex dependencies of a formal language, at a higher tier than any natural language, imposed by the grammar generator?

We investigate this question with OpenAI's frontier models GPT-5.5~\cite{GPT55SystemCard2026} and GPT-5.5-Pro~\cite{GPT55ProModel}, the latter exposing medium, high, and xhigh reasoning-effort configurations that scale test-time computation. We use ``reasoning'' and ``CoT'' as shorthand for this computation-level parameter, though the private computation of each model is neither observed or evaluated. These findings suggest a diagnostic inversion: more test-time reasoning has negligible effect of the already near-ceiling IXG test set, but substantially increases the acceptance of CSG-generated strings. Interestingly, while from the formal generator's perspective this may indicate sensitivity to higher-tier dependencies, from the perspective of English syntax, such acceptances are error: the more the models are allowed to ``reason'', the more we would consider their responses incorrect.
\textit{Are these ``mistakes'' evidence of formal language capabilities beyond that of human language rules?}

\section{Formal-language benchmark}
\label{sec:benchmark}

We reuse the 50 IXG  and 50 CSG strings selected by a previous study \cite{grahamgranger2026}, and defer formal definitions and generation details to Appendix~\ref{app:benchmark}. The two conditions are:
\begin{description}
    \item[IXG] strings with nested, serial, potentially long-range dependencies (including number agreement) of the sort seen in human languages. The generator is, specifically, a non-context-free tree-adjoining grammar (TAG) \cite{joshi1997tag}. TAG languages lie properly between the context-free and indexed languages \cite{ahoIndexedGrammars1968}, and are weakly equivalent to other mildly context-sensitive formalisms that were predominantly designed to capture non-context-free natural-language syntax \cite{shieber1985evidence,Steedman1987,Pollard1984,vijayshanker1994equivalence,ahoIndexedGrammars1968}.
    \item[CSG] strings generated with sequence-level, context-dependent agreement rules that may be beyond that of the IXG, exceeding the dependencies found in any human language rules. 
\end{description}

The strings contain English words, combine basic English rules, and are intentionally nonsensical. Their length were constrained to roughly 11-21 words (typical English sentence length), which restricts the power necessary to parse them but accounts for rejection based on statistically anomalous sentence length. We focus on the two upper tiers because judgments are near perfect acceptance for current large language models \cite{grahamgranger2026}. Each string is judged with the original binary prompt, inquiring for a ``yes''/``no'' acceptance of a sentence's syntax (detailed in Appendix \ref{app:benchmark}).

Language models are off-the-shelf, as no demonstrations, grammar definitions, or finetuning are provided. The evaluation tests zero-shot judgments of the deployed models we use today, not grammar induction from labeled examples.

Questions especially arise as a function of the difficulty of the data: are the most challenging sets of sentences actually being successfully parsed? In particular, were there any cases in which the system responded ``yes''  but was not in fact parsing the sentence? To further examine this issue, each model configuration is then prompted to to produce a parse of the sentence. In this way, whether a ``yes'' response aligns with the production of a parse whereas a ``no'' aligns with an attempted but failed parse, may be checked.

\section{GPT-5.5 and GPT-5.5-Pro evaluation}

\subsection{Models and design}
\label{sec:models}

We evaluated the April 23, 2026 snapshots \texttt{gpt-5.5-2026-04-23} (default) and \texttt{gpt-5.5-pro-2026-04-23} at medium, high, and xhigh reasoning effort, running  50 repeated trials over each 50-string set, i.e., 2{,}500 queries per model--grammar cell.
Standard deviations are across trials. Sampling parameters, API configuration, and cost accounting appear in Appendix~\ref{app:models}.

\subsection{Results}
\label{sec:results}

\subsubsection{Grammatical Judgment Responses}

\begin{figure}[tb]
    \centering
    \includegraphics[width=\linewidth]{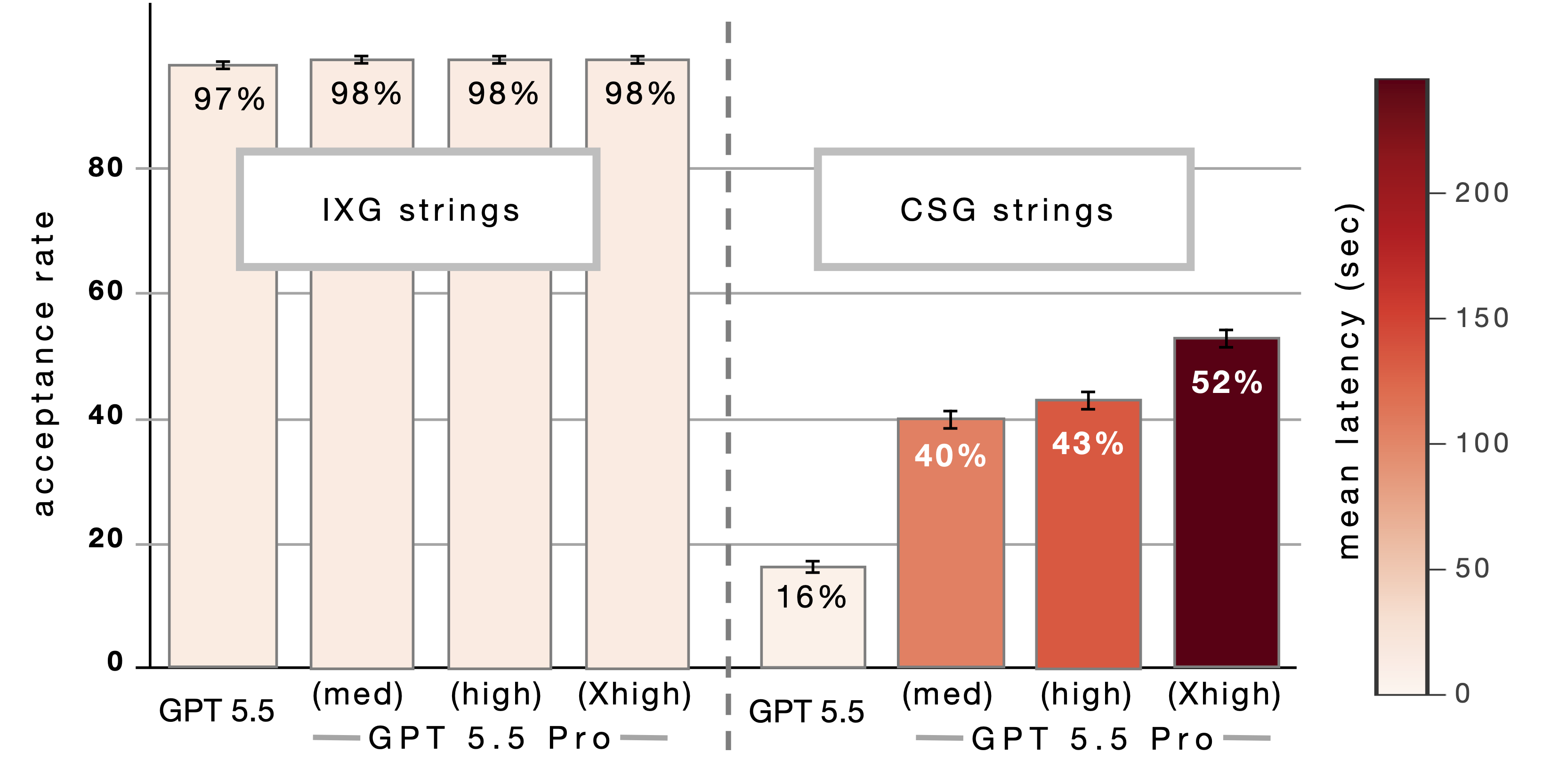}
    \caption{Acceptance rate by grammar generator and by reasoning-effort level, with color scale reflecting average elapsed query time. IXG-generated data acceptance is at ceiling. CSG acceptance scales monotonically with reasoning effort. Average query time (colors) also scales monotonically, both with grammar class and with reasoning effort.} 
     \label{fig:ch4:ii.results}
\end{figure}

Figure~\ref{fig:ch4:ii.results} plots the final results and Table~\ref{tab:results} reports the acceptance and query--time latency results.

\begin{table}[h]
    \centering
    \small
    \caption{Final results: means \(\pm\) SD across repeated trials over the 50-string sets.}
    \vspace{2mm}
    \label{tab:results}
    \renewcommand{\arraystretch}{1.15}
    \begin{tabular}{llccc}
    \toprule
        \bf Grammar & \bf Model/configuration & \bf Trials & \bf Acceptance (\%) & \bf Seconds/query \\
        \midrule
        \multirow{4}{*}{\textbf{IXG}}
        & GPT 5.5              & 50    & \(96.92 \pm 1.29\)  & \(3.87 \pm 0.98\)   \\
        & GPT 5.5 Pro (Med)    & 50    & \(97.84 \pm 0.55\)  & \(14.71 \pm 1.44\)   \\ 
        & GPT 5.5 Pro (High)   & 50    & \(97.91 \pm 0.57\)  & \(20.80 \pm 5.02\)    \\
        & GPT 5.5 Pro (XHigh)  & 50    & \(97.64 \pm 2.00\)  & \(50.33 \pm 25.53\)     \\
        \midrule
        \multirow{4}{*}{\textbf{CSG}}
        & GPT 5.5              & 50    & \(16.08 \pm 3.31\)  & \(7.10 \pm 2.27\)   \\
        & GPT 5.5 Pro (Med)    & 50    & \(40.35 \pm 4.64\)  & \(105.21 \pm 12.51\)   \\ 
        & GPT 5.5 Pro (High)   & 50    & \(42.76 \pm 4.41\)  & \(133.74 \pm 23.85\)    \\
        & GPT 5.5 Pro (XHigh)  & 50    & \(51.98 \pm 7.00\)  & \(241.00 \pm 75.54\)     \\         
    \bottomrule
    \end{tabular}
\end{table}

\paragraph{Pairwise acceptance tests.}
Table~\ref{tab:pairwise} reports paired tests on per-trial acceptance rates (difference being the second minus first configuration, in percentage points). All configurations were evaluated over 50 trials.

\begin{table}[h]
    \centering
    \vspace{5mm}
    \small
    \caption{Pairwise acceptance-rate comparisons. \(\Delta\) represents the percentage-point difference (second \(-\) first), \(p_t\) the paired \(t\)-test, \(p_W\) the Wilcoxon signed-rank, and \(d\) denotes Cohen's \(d\).}
    \vspace{2mm}
    \renewcommand{\arraystretch}{1.15}
    \begin{tabular}{llrrrr}
         \toprule
         \bf Grammar & \bf Comparison & \bf \centering \(\Delta\) & \bf \centering \(p_t\) & \centering \bf \(p_W\) & \bf \(d\) \\
         \midrule
         \multirow{6}{*}{\textbf{CSG}}
         & 5.5 \(\to\) Pro Med    & \(+24.27\) & \(1.6 \times 10^{-33}\) & \(7.3 \times 10^{-10}\) & \(-4.31\) \\
         & 5.5 \(\to\) Pro High   & \(+26.68\) & \(1.2 \times 10^{-37}\) & \(7.2 \times 10^{-10}\) & \(-5.28\) \\
         & 5.5 \(\to\) Pro XHigh  & \(+35.90\) & \(4.9 \times 10^{-34}\) & \(7.3 \times 10^{-10}\) & \(-4.42\) \\
         & Med \(\to\) High       & \(+2.41\) & \(1.1 \times 10^{-2}\) & \(1.4 \times 10^{-2}\)  & \(-0.37\) \\
         & Med \(\to\) XHigh      & \(+11.63\) & \(5.7 \times 10^{-13}\) & \(3.5 \times 10^{-9}\)  & \(-1.37\) \\
         & High \(\to\) XHigh     & \(+9.22\) & \(5.9 \times 10^{-10}\) & \(8.6 \times 10^{-9}\)  & \(-1.09\) \\
         \midrule
         \multirow{6}{*}{\textbf{IXG}}
         & 5.5 \(\to\) Pro Med    & \(+0.92\) & \(2.9 \times 10^{-5}\)    & \(1.0 \times 10^{-4}\) & \(-0.65\) \\
         & 5.5 \(\to\) Pro High   & \(+0.99\) & \(3.9 \times 10^{-6}\)    & \(3.6 \times 10^{-5}\)  & \(-0.74\) \\
         & 5.5 \(\to\) Pro XHigh  & \(+0.72\) & \(6.4 \times 10^{-2}\)    & \(7.1 \times 10^{-5}\)  & \(-0.27\) \\
         & Med \(\to\) High       & \(+0.07\) & \(0.460\)                 & \(0.470\)              & \(-0.11\) \\
         & Med \(\to\) XHigh      & \(-0.20\) & \(0.507\)                 & \(1.000\)               & \(0.09\) \\
         & High \(\to\) XHigh     & \(-0.27\) & \(0.362\)                 & \(0.608\)               & \(0.13\) \\
         \bottomrule
    \end{tabular}
    \label{tab:pairwise}
\end{table}

\paragraph{Latency.} All twelve pairwise latency differences are significant (Wilcoxon \(p \le 1.8 \times 10^{-15}\)), with very large effects for the baseline-vs-Pro comparison (e.g., CSG, GPT-5.5 vs Pro-medium: \(d=-7.72\)).

\paragraph{ANOVA.} Two-way factorial ANOVA on per-trial outcomes. Acceptance: model \(F= 460.06\), grammar \(F=26680.57\), interaction \(F=416.47\). Latency: model \(F=375.60\), grammar \(F=1109.29\), interaction \(F=167.12\). All significant with \(p < 10^{-16}\).

\subsubsection{Parses}

Both GPT-5.5 and GPT-5.5-Pro-xhigh were prompted with a follow-up query for a single trial of all 50 strings. Below are the models' parses, the \(G_{cs}\) grammar parses, and evaluations for two strings, one from each grammar.

The IXG-generated string ``\textit{Leibniz who invites Riemann who conjectures Galileo who rests solves the mountains that eat or fall}'' was accepted by both models. The parse for this string in terms of the grammar that generated it (specifically \(G_{ix}\) defined in the supplementary of material \cite{grahamgranger2026}) is provided for reference in Appendix~\ref{app:parses}, Figure~\ref{fig:G_ix-parse-string}.

\newpage
The following tree depicts the parse response from GPT-5.5:
\begin{center}
\begin{adjustbox}{max width=\linewidth}
\begin{forest} parse
[S
  [NP
    [N [Leibniz]]
    [CP [WHNP [who]]
      [S [VP [V [invites]]
        [NP [N [Riemann]]
          [CP [WHNP [who]]
            [S [VP [V [conjectures]]
              [NP [N [Galileo]]
                [CP [WHNP [who]]
                  [S [VP [V [rests]]]]
                ]
              ]
            ]]
          ]
        ]
      ]]
    ]
  ]
  [VP [V [solves]]
    [NP [Det [the]] [N [mountains]]
      [CP [WHNP [that]]
        [S [VP [VP [V [eat]]] [Conj [or]] [VP [V [fall]]]]]
      ]
    ]
  ]
]
\end{forest}
\end{adjustbox}
\end{center}

\newpage
The following tree depicts the parse response from GPT-5.5-Pro-xhigh:
\begin{center}
\begin{adjustbox}{max width=\linewidth}
\begin{forest} parse
[S
  [NP-SBJ
    [NNP [Leibniz]]
    [SBAR [WHNP [WP [who]]]
      [S [VP [VBZ [invites]]
        [NP [NNP [Riemann]]
          [SBAR [WHNP [WP [who]]]
            [S [VP [VBZ [conjectures]]
              [NP [NNP [Galileo]]
                [SBAR [WHNP [WP [who]]]
                  [S [VP [VBZ [rests]]]]
                ]
              ]
            ]]
          ]
        ]
      ]]
    ]
  ]
  [VP [VBZ [solves]]
    [NP [DT [the]] [NNS [mountains]]
      [SBAR [WHNP [WDT [that]]]
        [S [VP [VP [VBP [eat]]] [CC [or]] [VP [VBP [fall]]]]]
      ]
    ]
  ]
  [. [.]]
]
\end{forest}
\end{adjustbox}
\end{center}

The CSG-generated string ``\textit{the dogs that silently fade a person that deeply writes Laplace conjecture elegantly prove}'' was rejected by GPT-5.5 and received positive responses indicating grammatical acceptance by GPT-5.5-Pro-xhigh. The parse for this string in terms of the grammar that generated it (\(G_{cs}\) defined in the Supplementary Material of \cite{grahamgranger2026}) is provided for reference in Appendix~\ref{app:parses}, Figure \ref{fig:G_cs-parse-string}.

The following tree depicts the parse response from GPT-5.5, where ``ERR'' (\textit{red}) was the instructed symbol to mark any ``ill-formed or unattached material'':

\begin{center}
\begin{adjustbox}{max width=\linewidth}
\begin{forest} parse
[S
  [NP
    [Det [The]]
    [N [dogs]]
    [CP
      [C [that]]
      [S
        [VP
          [AdvP [silently]]
          [V [fade]]
          [NP
            [Det [a]]
            [N [person]]
            [CP
              [C [that]]
              [S
                [VP
                  [AdvP [deeply]]
                  [V [writes]]
                  [NP
                    [N [Laplace]]
                    [N [conjecture]]
                  ]
                ]
              ]
            ]
          ]
        ]
      ]
    ]
  ]
  [VP
    [AdvP [elegantly]]
    [V [prove]]
    [{\textcolor{red}{ERR}}]
  ]
  [. [.]]
]
\end{forest}
\end{adjustbox}
\end{center}

The following tree depicts the parse response from GPT-5.5-Pro-xhigh:

\begin{center}
\begin{adjustbox}{max width=\linewidth}
\begin{forest} parse
[S
  [NP
    [DT [The]]
    [NNS [dogs]]
    [SBAR
      [WHNP
        [WDT [that]]
      ]
      [S
        [VP
          [ADVP
            [RB [silently]]
          ]
          [VBP [fade]]
          [NP
            [DT [a]]
            [NN [person]]
            [SBAR
              [WHNP
                [WDT [that]]
              ]
              [S
                [VP
                  [ADVP
                    [RB [deeply]]
                  ]
                  [VBZ [writes]]
                  [NP
                    [NNP [Laplace]]
                    [NN [conjecture]]
                  ]
                ]
              ]
            ]
          ]
        ]
      ]
    ]
  ]
  [VP
    [ADVP
      [RB [elegantly]]
    ]
    [VBP [prove]]
  ]
  [. [.]]
]
\end{forest}
\end{adjustbox}
\end{center}

\subsection{Analysis}

Figure~\ref{fig:ch4:ii.results} and Table~\ref{tab:results} provide a plot of the final results and a detailed results report, respectively. IXG-generated acceptance stays between \(96.92\%\) and \(97.91\%\) averages for all models. Average CSG-generated string acceptance rises from \(16.08\%\) (GPT-5.5) to \(40.35\%\) (Pro, medium), \(42.76\%\) (Pro, high), and \(51.98\%\) (Pro, xhigh). A two-way model-by-grammar ANOVA on per-trial acceptance finds effects of the model (\(F(3,392)=460.06\)), grammar generator (\(F(1,392)=26680.57\)), and interaction (\(F(3,392)=416.47\)), all \(p< 10^{-16}\). This interaction shows that additional reasoning changes CSG-generated string acceptance far more than IXG string acceptance.

Paired comparison tells the same story. On CSG-generated strings, GPT-5.5 trails GPT-5.5-Pro-medium by 24.3 points (Wilcoxon \(p=7.3 \times 10^{-10}\)), GPT-5.5-Pro-xhigh exceeds medium by 11.6 points (\(p=3.5 \times 10^{-9}\)) and high by 9.2 points (\(p=8.6 \times 10^{-9}\)), and medium and high differ by 2.4 points (paired-\(t\) \(p=0.011\)). On IXG-generated strings, GPT-5.5 trailing the Pro configuration accepts strings at an increase of about 1 point, which is statistically detectable but behaviorally negligible at near total acceptance, while the Pro configurations are mutually indistinguishable (\(p>0.4\)). Latency scales steeply with effort (Table~\ref{tab:results}) and all pairwise latency differences are significant. Full test statistics, effect sizes, and ANOVA tables appear in Appendix~\ref{app:stats}.

The parse responses demonstrate that the models are to provide a parse of the strings. Whether or not such parsing informs the grammatical-acceptance response would require interpreting the internal model processing. What the parses help inform, interestingly, is that the models output parses similar in dependency depth and substructure relations. This is the case whether or not the model has responded positively about the given string.

\subsection{Higher-tier strings are not correct English}
\label{sec:example}

The over $50\%$ average acceptance of CSG-generated strings by GPT-5.5-Pro-xhigh is remarkable since the CSG-generated strings contain dependency relations that do not occur in any natural human language. Correspondingly,  human participants overwhelmingly judged these  strings to be nongrammatical. For example, here are CSG-generated strings that were consistently accepted in the Pro-xhigh trials:

\begin{quote}
    \it a brilliant bird with a lion and about the fish a soft person falls the complex fish
    \\ 

    the dogs that silently fade a person that deeply writes Laplace conjecture elegantly prove
    \\

    the brilliant equations from the intricate strawberries Turing silently go intricately read the dog who admires an ocean
\end{quote}
 
Human readers appropriately reject such strings as not grammatical  \cite{grahamgranger2026}. Increased LLM acceptance of such sentences could possibly reflect reasoning-induced rationalization of malformed input or shift in the decision criterion for answering ``yes'', but evidence (e.g., see Appendix C) indicates that these computationally powerful systems  (e.g., \cite{jiang2025softmaxtransformersturingcomplete}) are in fact inducing the parsing characteristics of the CSG generator.
 
Higher acceptance thus may not be
straightforwardly considered as ``better'' performance.  
These acceptances may reflect increased computational power, yet they also reveal misalignment:  the increased acceptance of these grammatically advanced strings is misaligned with the rejections given by human judgments.

\section{Discussion}

\paragraph{Does more reasoning enable a higher language?}
Additional test-time computation may let GPT-5.5-Pro track relations largely inaccessible to the baseline models. These increased computational capabilities are associated with higher classes of formal languages, as seen in the grammar-automata hierarchy.  
The evidence shows that test-time reasoning increases acceptance of strings from higher-tier generators.   (It always should be remembered in automata theory that any finite behavioral testing cannot itself identify a model's tier; this must be inferred from the {\em reliability} of test responses (with \emph{positive} and \emph{negative} samples) across extended corpora.)

\paragraph{Human grammaticality is the wrong ground truth here.}
As models increasingly process strings from grammars beyond the class containing studied natural language rules, the models may increasingly diverge from human alignment in judgments of grammaticality of strings composed of natural language symbols. 
Humans have been shown to accept CFG- and IXG-generated strings but reject CSG-generated strings, consistent with known natural-language dependencies \cite{grahamgranger2026, Steedman1987, Pollard1984, weirCombinatoryCategorialGrammars1988, joshi1997tag, ahoIndexedGrammars1968}. It follows that a model that accepts the CSG-generated strings here is misaligned with humans judgments, on one hand, but may be performing operations that are of computational interest and importance, on the other hand. 
The notion that advanced systems will be \emph{aligned} with human judgments is not a consistent concept, and benchmarks scored in accordance with human judgments or human alignment risk misreading computational capability gains as errors.

\paragraph{Limitations and the cost of stochastic evaluation.}
Additional limitations qualify these results. First, benchmarks can only evaluate systems on a finite set of measures: they can provide evidence for membership in a particular computational class, but cannot provide decisive proof. Furthermore, the test data was highly limited by specific string-selection criteria for greater English alignment. Additionally, language foundation models are prompt-sensitive and the prompt wording may affect the responses and may conflate structural recognition with instruction following and response calibration.
Thirdly, we queried only $50$ times, at the lower end of the developer stochasticity range (temperature $0.2$) which determines the determinism of a model's output. Fifty trials-per-query is already expensive: the present partial matrix comprises $20{,}000$ API calls, $\approx 450$ serial query-hours including small wait time between queries, and $\$5{,}000$ at list prices, as reasoning tokens are billed as output tokens \cite{PricingOpenAIAPI} (see Appendix~\ref{app:models}). 
Note that binary responses, though  more cost-efficient, limits overall understanding of the model responses.

\section{Conclusion}

Increasing GPT-5.5-Pro's test-time reasoning effort was found to monotonically raise its acceptance of grammatical constructs from a context-sensitive grammar generator.
This is evidence that added computation may improve a model's computational ability to process complex relational dependencies in data. 
In sum,
as reasoning inference-time computations increase, language model performance shows evidence of a monotonically increasing ability to successfully process complex data structures. 

But it also highlights  distinctions between high-level reasoning, on one hand, and alignment with human judgments, on another hand. As GPT-5.5-Pro's test-time compute increases, it accepts the grammaticality of candidate sentences that humans would (rightfully) reject, since they are not grammatical sentences in English (nor any human natural language).  

For instance, ``the dogs that silently fade a person that deeply writes Laplace conjecture elegantly prove,'' is a string generated by a context sensitive grammar; this string and others at this tier were judged acceptable by GPT-5.5-Pro, and yet judged ungrammatical by human subjects.  
Notably, both responses may be considered correct,  depending on the metric: GPT-5.5-Pro models showed evidence (see sample parses above and in Appendix C) of parsing strings from a context sensitive grammar generator, whereas humans (also correctly) recognized that the string did not conform to any English grammar rules.
While humans are using  language processing abilities based on their knowledge of their native tongue, AI models acquire related knowledge through extensive training on large natural language text corpora.  But advanced AI models may also be able to discern patterns beyond those that occur in the natural language rules formalized and studied by computational linguists, and thus may be able to reliably process some data from generators at higher levels such as CSG.

In other words, more `reasoning'-level computation  enables models to accept string constructions that contain dependencies beyond those found in human natural languages.
To evaluate a model, it  is therefore necessary to distinguish between
fundamental computational ability on one hand, and alignment with human judgments on the other. 
These results offer initial quantitative evidence of a crucial insight: improved ability is not always improved alignment.

\begin{ack}
{ This work was supported in part by funds from the Office of Naval Research.}
\end{ack}

{\small
\bibliographystyle{unsrt}
\bibliography{references}
}

\newpage
\appendix
\section{Benchmark details}
\label{app:benchmark}

\paragraph{Formal preliminaries.}
A formal grammar is a tuple $G = (V, \Sigma, R, S)$, where $V$ is a finite set of nonterminal symbols, $\Sigma$ is a finite set of terminal symbols, $R$ a set of production rules, and $S\in V$ the start symbol. A formal language $\mathcal{L}(G)$ is the set of all strings derivable from $S$. The grammar--automata (abstract machine) hierarchy (Figure~\ref{fig:hierarchy}) defines classes of formal grammars and automata that are equivalent in terms of computational power, specifically the formal languages that they can generate and recognize, respectively. It is worth noting in more detail here that indexed grammar class properly contains the context-free grammars, the tree-adjoining and mildly context-sensitive grammars, and are properly contained in the context-sensitive grammar class \cite{ahoIndexedGrammars1968,weirCharacterizingMildly1988}.

\paragraph{Test data generation and selection.}
We reuse the 50 IXG and 50 CSG strings from Graham and Granger \cite{grahamgranger2026}, which are publicly available: \url{https://github.com/DartmouthGrangerLab/ThreeTiers}. The grammars use English words as terminals, combine basic English syntactic rules (e.g., noun phrase, then verb phrase; determiner then noun), and are intentionally nonsensical, so that judgments cannot be based on memorized sentences or semantic plausibility. Selected strings were limited to approximately 20--21 words. The IXG is, specifically, a tree-adjoining grammar (TAG), and the CSG implements sequence-level, context-dependent agreement exceeding the dependencies observed in English. The original study reports that human participants accept CFG- and IXG-generated strings and overwhelmingly reject the CSG-generated strings.

\paragraph{Querying}
The prompt from the original study was reused:
\begin{quote}
    \it
    Disregarding any lack of punctuation or sensicality, please answer the following with only ``Yes'' or ``No'': If the following sentence grammatically complete?
\end{quote}
The string being tested was appended to the prompt to form the full query seen by the models. The wording ``grammatically complete'' was adopted by the original study instead of ``grammatically correct'' because the smallest evaluated models demonstrated issues separating syntactic well-formedness with commonsense validity (a consequential finding of their inability to follow the ``yes'' and ``no'' responses only prompt) \cite{grahamgranger2026}.

The follow-up query requesting a parse of the string was:
\begin{quote}
        \it
        Parse the sentence exactly as written. Use a bracketed format such as:
        \begin{center}
            \([\text{S}\; [\text{NP} \,\dots] \, [\text{VP} \,\dots]\,]\)
        \end{center}
        Return only the bracketed parse.
\end{quote}

The follow-up query was accompanied by the following input into the \texttt{instructions} GPT 5.5 (Pro) configuration:
\begin{quote}
    \it Parse the sentence exactly as written without correcting, rewriting, or normalizing it. The grammaticality was determined by your previous response. If it is ungrammatical, provide the best bracketed parse possible and mark any ill-formed or unattached material with an ERR label. Return only the bracketed parse, without Markdown, explanation, or commentary.
\end{quote}

\newpage
\section{Models and API Configuration}
\label{app:models}

\paragraph{Configurations.} Model: \texttt{gpt-5.5-2026-04-23} (default effort) and \\\texttt{gpt-5.5-pro-2026-04-23} with reasoning effort \texttt{medium}, \texttt{high}, \texttt{xhigh} (\texttt{medium} is the API default for Pro). Sampling parameters: \texttt{temperature} \(=0.2\), \texttt{verbosity} \(=\) \texttt{low}. System message: \emph{``Answer only yes or no.''} The API returns only the final Yes/No answer. Intermediate reasoning is not exposed. 

\paragraph{Data Audit.} A comprehensive data audit confirmed that all expected data points across the 20,000 queries are fully populated and valid, with no missing trials.

\paragraph{Query volume, time calculations, and cost.} Testing 50 trials over 50 prompts for each of the two grammars across all four model configurations (GPT-5.5, Pro-Medium, Pro-High, Pro-XHigh), the computation for the final results is \(4 \times 50 \times 50 \times 2 = 20{,}000\) total queries. 

Because not all trials successfully recorded execution time, the average active query time was calculated per question ``id'' over all available time information in every trial for each model. The total active time was then estimated by summing these per-question averages and multiplying by the number of trials (50) for each grammar. To avoid exceeding API rate limitations, a latent wait time of 10 seconds was intentionally introduced between queries. 

Summing the estimated active model execution time (\(\approx 16\) days, 16 hours, 51 minutes) and the latent wait time (\(20{,}000 \times 10\) seconds \(\approx 2\) days, 7 hours, 33 minutes) yields a grand total runtime of approximately 19 days, 0 hours, and 25 minutes (or \(\approx 456\) serial query-hours). The detailed breakdown of these runtime estimates is provided in Table~\ref{tab:runtime_summary}.

At evaluation time, GPT-5.5 was priced at \(\$5\)/\(\$30\) per million input/output tokens and GPT-5.5-Pro at \(\$30\)/\(\$180\) \cite{PricingOpenAIAPI} (reasoning tokens are billed as output). The observed total expenditure for these completed runs was \(\approx \$5{,}000\). Fifty repetitions yield \(2{,}500\) observations per completed cell.

\begin{table}[ht]
    \centering
    \small
    \caption{Estimated Runtime Summary (50 Trials + 10s Wait Latency Per Query)}
    \label{tab:runtime_summary}
    \renewcommand{\arraystretch}{1.15}
    \resizebox{\textwidth}{!}{
    \begin{tabular}{lccrrrr}
        \toprule
        \bf Model & \bf Grammar & \bf Strings & \bf Avg Query (s) & \bf Active (50 trials) & \bf Wait (10s/q) & \bf Total Estimated \\
        \midrule
        GPT 5.5 & IXG & 50 & \(3.87\) & 2h 41m 06.7s & 6h 56m 40.0s & 9h 37m 46.7s \\
        GPT 5.5 & CSG & 50 & \(7.10\) & 4h 55m 56.7s & 6h 56m 40.0s & 11h 52m 36.7s \\
        GPT 5.5 Pro (Med) & IXG & 50 & \(14.71\) & 10h 12m 48.7s & 6h 56m 40.0s & 17h 09m 28.7s \\
        GPT 5.5 Pro (Med) & CSG & 50 & \(105.24\) & 3d 01h 04m 52.0s & 6h 56m 40.0s & 3d 08h 01m 32.0s \\
        GPT 5.5 Pro (High) & IXG & 50 & \(20.80\) & 14h 26m 40.6s & 6h 56m 40.0s & 21h 23m 20.6s \\
        GPT 5.5 Pro (High) & CSG & 50 & \(133.74\) & 3d 20h 52m 38.1s & 6h 56m 40.0s & 4d 03h 49m 18.1s \\
        GPT 5.5 Pro (XHigh) & IXG & 50 & \(50.66\) & 1d 11h 10m 44.4s & 6h 56m 40.0s & 1d 18h 07m 24.4s \\
        GPT 5.5 Pro (XHigh) & CSG & 50 & \(241.13\) & 6d 23h 26m 54.8s & 6h 56m 40.0s & 7d 06h 23m 34.8s \\
        \bottomrule
    \end{tabular}
    }
\end{table}

\newpage
\section{A Parse in Terms of the Generating Grammar}
\label{app:parses}

\begin{figure}[h]
    \centering
    \includegraphics[width=\linewidth]{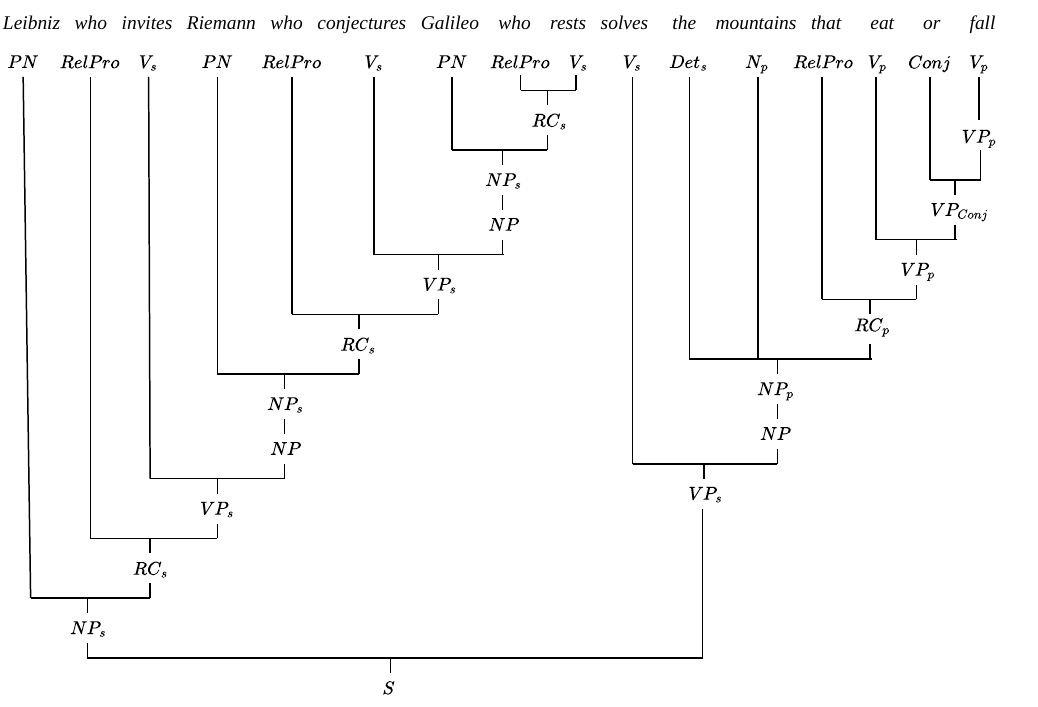}
    \caption{A valid parse in terms of the grammar generator \(G_{ix}\) of its string generation: \\ \textit{Leibniz who invites Riemann who conjectures Galileo who rests solves the mountains that eat or fall.}}
    \label{fig:G_ix-parse-string}
\end{figure}

Figure~\ref{fig:G_ix-parse-string} depicts a valid parse of the string \emph{Leibniz who invites Riemann who conjectures Galileo who rests solves the mountains that eat or fall}, in terms of the grammar that generated it: \(G_{ix}\) from \cite{grahamgranger2026}.

\vspace{5mm}

\newpage

\begin{figure}[h]
    \centering
    \includegraphics[width=\linewidth]{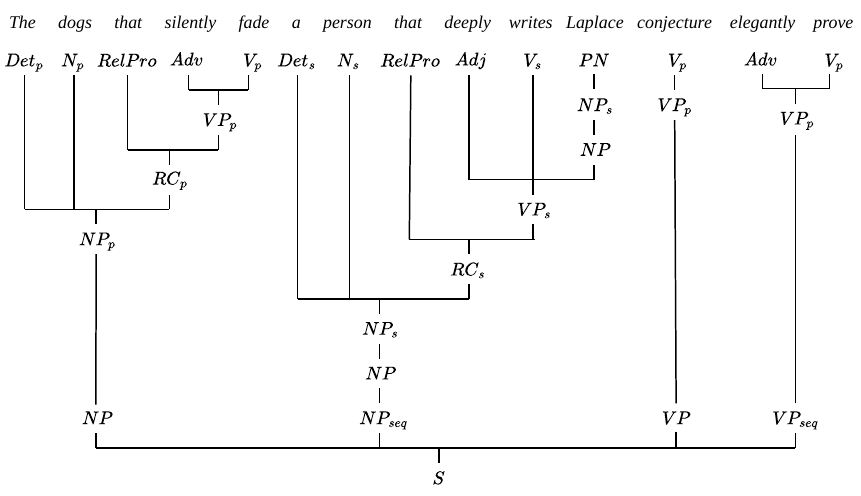}
    \caption{A valid parse in terms of the grammar generator \(G_{cs}\) of its string generation: \\ \textit{The dogs that silently fade a person that deeply writes Laplace conjecture elegantly prove.}}
    \label{fig:G_cs-parse-string}
\end{figure}

Figure~\ref{fig:G_cs-parse-string} depicts a valid parse of the string \emph{the dogs that silently fade a person that deeply writes Laplace conjecture elegantly prove}, in terms of the grammar that generated it: \(G_{cs}\) from \cite{grahamgranger2026}.

\vspace{5mm}

\clearpage
\section{Extended Statistical Analysis}
\label{app:stats}

While the main text details the acceptance rate means and pairwise comparisons, this section provides the extended statistical tables for query latency differences and the full Two-Way Factorial ANOVA results.

\paragraph{Query Latency Pairwise Tests.}
Table~\ref{tab:latency_pairwise} reports paired tests on per-trial query latency (difference = second minus first configuration, in seconds). All twelve pairwise latency differences are statistically significant.

\begin{table}[h]
    \centering
    \small
    \caption{Pairwise Query Latency comparisons. \(\Delta\) represents the mean difference in seconds (second \(-\) first), \(p_t\) the paired \(t\)-test, \(p_W\) the  Wilcoxon signed-rank, and \(d\) denotes Cohen's \(d\).}
    \vspace{2mm}
    \renewcommand{\arraystretch}{1.15}
    \begin{tabular}{llrrrr}
         \toprule
         \bf Grammar & \bf Comparison & \bf \centering \(\Delta\) (s) & \bf \centering \(p_t\) & \centering \bf \(p_W\) & \bf \(d\) \\
         \midrule
         \multirow{6}{*}{\textbf{CSG}}
         & 5.5 \(\to\) Pro Med    & \(+98.11\) & \(1.5 \times 10^{-45}\) & \(1.8 \times 10^{-15}\) & \(-7.72\) \\
         & 5.5 \(\to\) Pro High   & \(+126.64\) & \(2.2 \times 10^{-37}\) & \(1.8 \times 10^{-15}\)  & \(-5.21\) \\
         & 5.5 \(\to\) Pro XHigh  & \(+233.90\) & \(6.2 \times 10^{-27}\) & \(1.8 \times 10^{-15}\)  & \(-3.09\) \\
         & Med \(\to\) High       & \(+28.53\) & \(1.7 \times 10^{-9}\) & \(1.1 \times 10^{-7}\)  & \(-1.04\) \\
         & Med \(\to\) XHigh      & \(+135.79\) & \(1.6 \times 10^{-17}\) & \(3.6 \times 10^{-15}\)  & \(-1.84\) \\
         & High \(\to\) XHigh     & \(+107.26\) & \(5.1 \times 10^{-13}\) & \(3.6 \times 10^{-15}\)  & \(-1.37\) \\
         \midrule
         \multirow{6}{*}{\textbf{IXG}}
         & 5.5 \(\to\) Pro Med    & \(+10.84\) & \(2.8 \times 10^{-42}\)    & \(1.8 \times 10^{-15}\) & \(-6.60\) \\
         & 5.5 \(\to\) Pro High   & \(+16.93\) & \(2.3 \times 10^{-28}\)    & \(1.8 \times 10^{-15}\)  & \(-3.33\) \\
         & 5.5 \(\to\) Pro XHigh  & \(+46.47\) & \(4.1 \times 10^{-17}\)    & \(1.8 \times 10^{-15}\)  & \(-1.80\) \\
         & Med \(\to\) High       & \(+6.09\) & \(8.2 \times 10^{-11}\)                 & \(2.0 \times 10^{-9}\)              & \(-1.17\) \\
         & Med \(\to\) XHigh      & \(+35.63\) & \(3.6 \times 10^{-13}\)                 & \(1.8 \times 10^{-15}\)               & \(-1.39\) \\
         & High \(\to\) XHigh     & \(+29.53\) & \(2.4 \times 10^{-10}\)                 & \(4.4 \times 10^{-14}\)               & \(-1.12\) \\
         \bottomrule
    \end{tabular}
    \label{tab:latency_pairwise}
\end{table}

\paragraph{Two-Way Factorial ANOVA.} 
Tables~\ref{tab:anova_acceptance} and \ref{tab:anova_latency} present the two-way factorial ANOVA (Model \(\times\) Grammar Interaction) for Acceptance Rate and Query Latency, respectively. All main effects and interactions are significant with \(p < 10^{-16}\) (reported as \(0.0\)).

\begin{table}[h]
    \centering
    \small
    \caption{Two-Way Factorial ANOVA: Acceptance Rate}
    \vspace{2mm}
    \renewcommand{\arraystretch}{1.15}
    \begin{tabular}{lrrrr}
         \toprule
         \bf Source & \bf Sum Sq & \bf df & \bf \(F\) & \bf \(PR(>F)\) \\
         \midrule
         Model & \(1.84903\) & \(3.0\) & \(460.056\) & \(0.0\) \\
         Grammar & \(35.74426\) & \(1.0\) & \(26680.572\) & \(0.0\) \\
         Model \(\times\) Grammar & \(1.67384\) & \(3.0\) & \(416.467\) & \(0.0\) \\
         Residual & \(0.52517\) & \(392.0\) & -- & -- \\
         \bottomrule
    \end{tabular}
    \label{tab:anova_acceptance}
\end{table}

\begin{table}[h]
    \centering
    \small
    \caption{Two-Way Factorial ANOVA: Query Latency}
    \vspace{2mm}
    \renewcommand{\arraystretch}{1.15}
    \begin{tabular}{lrrrr}
         \toprule
         \bf Source & \bf Sum Sq & \bf df & \bf \(F\) & \bf \(PR(>F)\) \\
         \midrule
         Model & \(1.002376 \times 10^6\) & \(3.0\) & \(375.600\) & \(0.0\) \\
         Grammar & \(9.868030 \times 10^5\) & \(1.0\) & \(1109.295\) & \(0.0\) \\
         Model \(\times\) Grammar & \(4.459899 \times 10^5\) & \(3.0\) & \(167.117\) & \(0.0\) \\
         Residual & \(3.487141 \times 10^5\) & \(392.0\) & -- & -- \\
         \bottomrule
    \end{tabular}
    \label{tab:anova_latency}
\end{table}


\end{document}